\documentclass[10pt,twocolumn,letterpaper]{article}

\usepackage{cvpr}              
\definecolor{cvprblue}{rgb}{0.21,0.49,0.74}
\usepackage[pagebackref,breaklinks,colorlinks,allcolors=cvprblue]{hyperref}

\def\paperID{*****} 
\def\confName{CVPR}
\def\confYear{2026}

\title{ProtoAnomalyNCD: Prototype Learning for Multi-class Novel Anomaly Discovery in Industrial Scenarios}

\author{
Botong Zhao \quad
Qijun Shi \quad
Shujing Lyu \quad
Yue Lu\thanks{*Corresponding author.}\\
Shanghai Key Laboratory of Multidimensional Information Processing, East China Normal University\\
{\tt\small ylu@cee.ecnu.edu.cn}
}

\begin{document}
\maketitle
\begin{abstract}
Existing industrial anomaly detection methods mainly determine whether an anomaly is present. However, real-world applications also require discovering and classifying multiple anomaly types. Since industrial anomalies are semantically subtle and current methods do not sufficiently exploit image priors, direct clustering approaches often perform poorly.
To address these challenges, we propose ProtoAnomalyNCD, a prototype-learning-based framework for discovering unseen anomaly classes of multiple types that can be integrated with various anomaly detection methods. First, to suppress background clutter, we leverage Grounded SAM with text prompts to localize object regions as priors for the anomaly classification network. Next, because anomalies usually appear as subtle and fine-grained patterns on the product, we introduce an Anomaly-Map-Guided Attention block. Within this block, we design a Region Guidance Factor that helps the attention module distinguish among background, object regions, and anomalous regions. By using both localized product regions and anomaly maps as priors, the module enhances anomalous features while suppressing background noise and preserving normal features for contrastive learning. Finally, under a unified prototype-learning framework,  ProtoAnomalyNCD discovers and clusters unseen anomaly classes while simultaneously enabling multi-type anomaly classification. We further extend our method to detect unseen outliers, achieving task-level unification. Our method outperforms state-of-the-art approaches on the MVTec AD, MTD, and Real-IAD datasets.
\end{abstract}    
\section{Introduction}
\label{sec:intro}

Industrial anomaly detection has recently made remarkable progress~\cite{ma2025aa, luo2025exploring, gu2024filo, he2024mambaad}. 
However, most existing methods remain essentially binary: they can localize anomalous regions on products but cannot reliably recognize fine-grained anomaly categories or cope with the continual emergence of unseen types and rare cases.

\begin{figure*}[htb]
        \centering
	\includegraphics[width=0.74\textwidth]{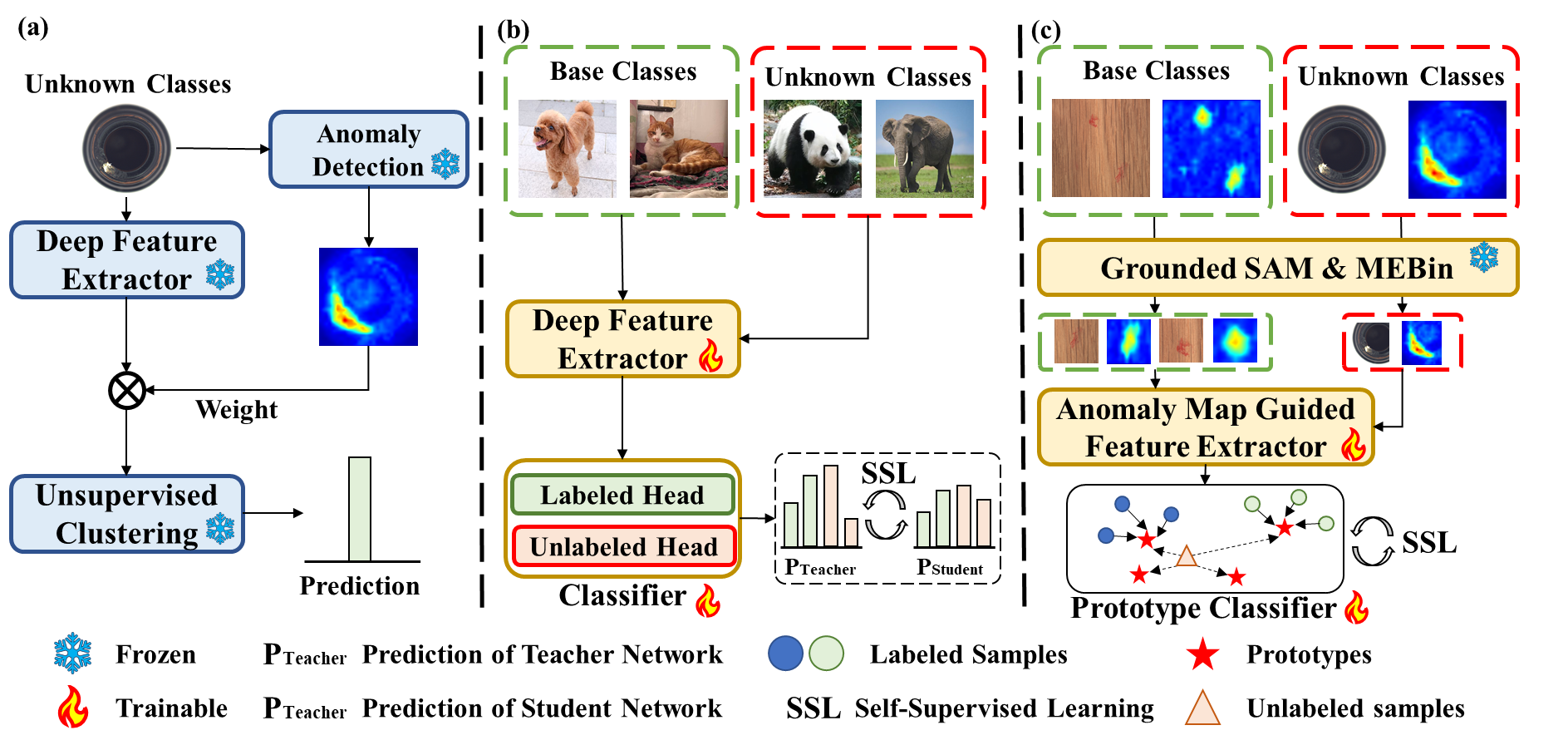}
	\caption{ Comparison between solutions organizing anomalies into groups. (a) Clustering-based methods extract features from anomaly regions and apply unsupervised clustering. (b) Vanilla NCD methods use a trainable feature extractor and classifier on object-centered images from both known and unknown classes. (c) Our ProtoAnomalyNCD learns anomaly prototypes directly from anomaly-centered crops and masks to perform classification.}
	\label{Fig:1}       
\end{figure*}

As shown in Fig.~\ref{Fig:1}(a)~\cite{sohn2023anomaly, lee2026uniformaly, ardelean2024blind}, clustering-based methods for multi-type anomaly classification typically follow two steps: first detecting anomalous regions, and then clustering features extracted from these regions. However, when anomalies share similar shapes, appearances, or spatial locations, performance degrades. This motivates us to exploit both intrinsic cues in anomaly images and known anomalies as joint priors to classify unknown anomalies.

Our analysis reveals three key challenges for classification networks in industrial anomaly scenarios.
1) Models trained on natural images usually assume a single, centered, independent object, whereas industrial anomalies appear as local patterns on the object itself.  
2) Industrial anomalies exhibit diverse shapes and weak semantic cues.  
3) Owing to the inherent randomness of anomalies, the number of anomaly categories cannot be predetermined.

In this work, we overcome the aforementioned challenges and introduce ProtoAnomalyNCD, a prototype-learning-based self-supervised framework for multi-type anomaly classification that aligns with the concept of Novel Class Discovery (NCD), as shown in Fig. 1(c). By directing the model’s attention to true anomalous regions, ProtoAnomalyNCD enables the discovery and classification of novel anomaly types and further extends to unseen out-of-distribution anomaly detection, achieving a unified treatment of both tasks.

To focus on the inspected object and capture the relationship between its normal and anomalous regions, we employ Grounded SAM with text prompts to localize the object regions and propose an Anomaly-Map-Guided Attention block. Leveraging anomaly maps as priors, this block enhances anomalous features while preserving the object’s semantic information. During training, we represent each anomaly type with a prototype to model relationships across anomaly categories, and we adopt corrected pseudo-labels to prevent false positives from contaminating the learning process. Finally, we introduce a criterion for estimating the number of unseen anomaly classes by jointly analyzing the feature space and the classification performance of known categories, enabling accurate identification of new anomaly types even when their number is unknown. Experiments on the MVTec AD, MTD, and Real-IAD datasets demonstrate the effectiveness of our method.

The main contributions are summarized as follows.
1)We propose ProtoAnomalyNCD, a prototype-learning-based self-supervised framework for unseen anomaly class discovery in industrial inspection, which can be seamlessly combined with various anomaly detection methods and categorizes previously unknown anomalies.
2) Our prototype-learning framework exploits priors from anomaly images and anomaly maps, using Grounded SAM to guide the model toward objects and their anomaly regions while modeling their relationships.
3) Extensive experiments on the MVTec AD, MTD, and Real-IAD datasets show that ProtoAnomalyNCD consistently outperforms existing anomaly clustering and NCD methods, providing a strong basis for downstream applications.

\section{Related Works}
\label{sec:Related Works}
\subsection{Industrial Anomaly Detection}
Industrial anomaly detection has evolved from single-class inspection to few-shot and multi-class settings. Early single-class methods based on reconstruction~\cite{cao2025varad, luo2024ami, zhang2024realnet}, knowledge distillation~\cite{tien2023revisiting, deng2022anomaly}, or embedding-based modeling~\cite{liu2023simplenet, zhang2023destseg} are typically tailored to specific products, limiting scalability across diverse categories. Few-shot approaches improve generalization via spatial alignment~\cite{huang2022registration} or contrastive learning~\cite{jiang2024prototypical}. More recently, vision–language models  have shown strong performance by leveraging rich pre-trained knowledge, as demonstrated in WinCLIP~\cite{jeong2023winclip}, AnomalyGPT~\cite{gu2024anomalygpt}, and AA-CLIP~\cite{ma2025aa}.

Despite this progress, most methods are still binary, separating only normal from abnormal, and naive clustering performs poorly for multi-class anomaly classification. To handle fine-grained anomaly types, recent work designs clustering pipelines using weighted patch aggregation~\cite{sohn2023anomaly}, visual–textual feature alignment~\cite{sadikaj2025multiads}, or high-score patch selection~\cite{lee2026uniformaly}. Huang et al.~\cite{huang2025anomalyncd} further adopt self-supervised learning to classify unlabeled anomalous regions. However, these pipelines rely on frozen feature extractors and assume a predefined number of categories.

To overcome these limitations, we adopt Prototype Learning to discover new anomaly categories by constructing pseudo-labels during prototype exploration, enabling classification from unlabeled anomalous regions and detecting isolated outlier samples.

\subsection{Novel Class Discovery}
Novel Class Discovery (NCD) was originally formulated as a deep transfer clustering problem~\cite{han2019learning}, where knowledge from labeled classes is used to cluster unlabeled data from unknown categories. Typical methods perform self-supervised pretraining, fine-tune on labeled data, and then learn novel classes using pseudo-labeling or interactions between labeled and unlabeled data~\cite{fini2021unified, caron2020unsupervised, zhong2021openmix, zhong2021neighborhood}. However, NCD usually requires prior knowledge of the number of categories in the unlabeled data, which limits its applicability in industrial anomaly detection scenarios. We instead develop a unified modeling approach for both base and novel classes: by introducing learnable prototypes, we design a pseudo-labeling mechanism that mitigates confirmation bias and estimates the number of novel classes from relationships between prototypes of base and novel classes.

\subsection{Prototype Learning}
Prototype learning~\cite{snell2017prototypical} extracts representative prototypes from training data and classifies test samples by their distances to these prototypes. It is widely used in few-shot learning~\cite{li2021adaptive} and has also been adopted in anomaly detection~\cite{gong2019memorizing, huang2022pixel, lv2021learning, park2020learning}. For example, PatchCore~\cite{roth2022towards} represents normal patterns with prototypes and detects anomalies via nearest distances. However, such methods only determine whether anomalies exist. In this work, we refine pseudo-labels using anomaly scores and learnable region-level anomaly prototypes, enabling the discovery of multiple anomaly categories under weak semantic cues.

\section{Proposed Method}
\label{sec:Proposed}

\begin{figure*}[htb]
        \centering
	\includegraphics[width=0.74\textwidth]{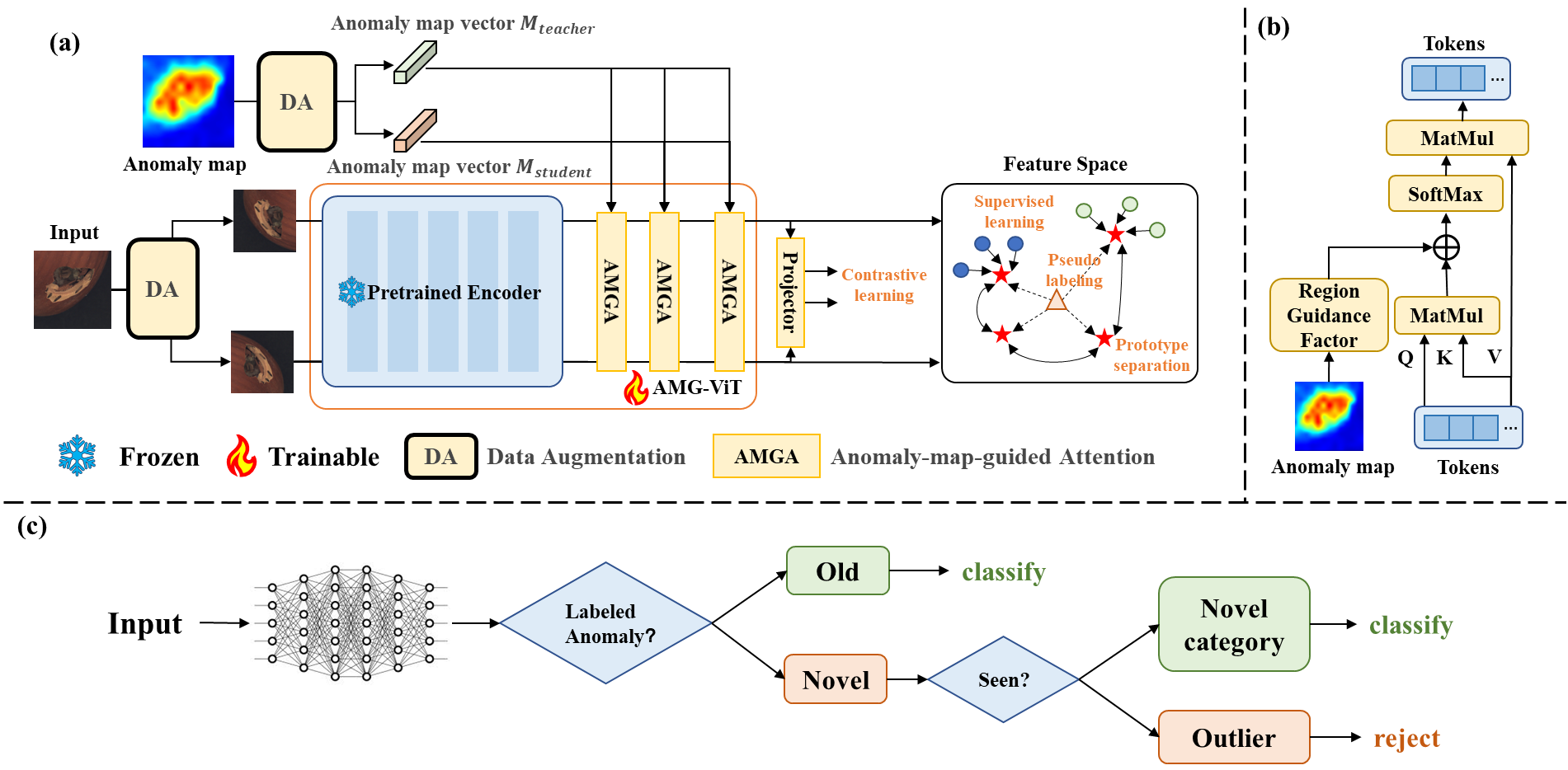}
	\caption{(a) Overview of the ProtoAnomalyNCD training pipeline.
(b) Structure of the anomaly-map-guided attention module.
(c) During inference, ProtoAnomalyNCD can classify both previously known and novel classes, and can also be extended to reject unseen outliers.}
	\label{Fig:2}       
\end{figure*}

ProtoAnomalyNCD aims to automatically discover and classify visual anomaly categories in industrial inspection. As illustrated in Fig. 2(a), we first extract the main object regions by applying Grounded SAM to separate foreground from background, obtain dominant anomaly areas via principal-element binarization, and feed the resulting anomaly map as prior knowledge into the anomaly-map-guided ViT (Sec. 3.2). Using these regions, we then perform anomaly-map-guided prototype learning (Sec. 3.3) to derive prototypes for different anomaly types and obtain discriminative features for classification. Finally, we estimate the number of unseen anomaly categories by analyzing both the feature space and the base classes (Sec. 3.4). As illustrated in Fig. 2(c), We further extend ProtoAnomalyNCD to out-of-distribution detection (Sec. 3.5), enabling the identification of object-irrelevant outlier samples.

\subsection{Problem Definition}

Given a set of unlabeled anomalous images
\[
\mathcal{D}^u = \{ I_i^u \mid i \in [1, N^u] \},
\]
the goal of ProtoAnomalyNCD is to discover prototypes of the \( C^u \) unknown categories (novel classes) present in these images and assign each sample to one of them.

To effectively learn novel classes, we follow the common NCD setting and assume the existence of a labeled anomalous dataset
\[
\mathcal{D}^l = \{ (I_i^l, y_i^l, M_i^l) \mid i \in [1, N^l] \},
\]
which contains \( C^l \) known categories (base classes). Here, \( y_i^l \in \mathbb{R}^{1 \times (C^l + C^u)} \) denotes the one-hot label of image \( I_i^l \), and \( M_i^l \) denotes its anomaly map. This labeled dataset provides prior knowledge to improve the clustering ability on the unlabeled set \( \mathcal{D}^u \).

Existing work \cite{huang2025anomalyncd, sohn2023anomaly} often assumes that \( C^u \) is known. In contrast, we argue that \( C^u \) is unknown in realistic industrial open-set scenarios, and propose an estimation method for \( K_{\text{new}} \) in Sec.~3.4. Therefore, the total number of categories is $K = K_{\text{base}} + K_{\text{new}} + 1,$where the additional 1 corresponds to the normal class.

Let \( \mathcal{E}(\cdot) \) denote the feature extractor and \( \phi(\cdot) \) denote the projection head. For a sample \( x_i \), its feature representation is$z_i = \mathcal{E}(x_i),$ and its projected representation in the contrastive space is $h_i = \phi(z_i),$ where \( z_i \in \mathbb{R}^{d} \) and \( h_i \in \mathbb{R}^{d_h} \).

To evaluate performance under different conditions, we conduct experiments using either both \( \mathcal{D}^l \cup \mathcal{D}^u \) or only \( \mathcal{D}^u \).

\subsection{Anomaly-Map-Guided ViT}
Industrial anomaly patterns differ fundamentally from natural-image objects: rather than appearing as independent entities located near the center of the image, industrial anomalies manifest as subtle and fine-grained local deviations on the object surface. These weak signals are easily overwhelmed by background textures, and the object itself often carries strong correlations with normal patterns. To better isolate these fine-grained anomalies, we first generate an anomaly map $A_i$ using INP-Former and then segment the object region via Grounded SAM with text prompts. The resulting foreground mask is further stabilized using MEBin binarization, which adaptively selects thresholds to extract the major structural components of the anomaly region while preserving the spatial layout of the object.

The proposed anomaly-map-guided ViT (AMG-ViT) builds on the standard Vision Transformer\cite{dosovitskiy2020image}.
However, the strong preference of ViT for global object structure often causes fine anomalies to be overlooked, leading to insufficient attention on the regions where anomaly occur.

The anomaly map is used to guide the attention mechanism. The image is first split into $N$ patches, and the anomaly map is downsampled by average pooling to produce a vector aligned with the patch tokens. This anomaly vector is then injected into the attention computation, encouraging the model to focus on anomalous regions:

\[
\mathrm{Attn}
=\mathrm{softmax}\!\left(
[Q^{cls}K^\top+\mathcal{M},\; Q^{p}K^\top]
\right)V .
\]

where $\mathcal{M}$ denotes the anomaly-map vector.

To ensure compatibility with the attention mechanism, we design a Region Guidance Factor:

\[
M(i)=
\begin{cases}
0, & 0 \le i < \tau_1, \\
\gamma \log\left(\frac{d(i)}{\tau}\right), & \tau_1 \le i < \tau_2, \\
-\infty, & i \ge \tau_2,
\end{cases}
\]
where $i$ is the anomaly score, $\tau_1$ and $\tau_2$ define low- and high-confidence anomaly regions, and $\gamma$ controls the smoothness of the weighting transition.

With the anomaly-guided adjustment applied to the final Transformer layer, AMG-ViT consistently directs more attention toward anomalous regions. 

\subsection{ProtoAnomalyNCD}
\subsubsection{Prototype-Based Probabilistic Modeling}
To unify the representation of base and novel classes, we model all categories within a shared hyperspherical feature space by normalizing their \( d \)-dimensional representations onto the unit sphere \( \mathbb{S}^{d-1} \).  
This shared space enables the model to leverage label-supervised knowledge while reducing class bias \cite{chen2020simple, wang2020understanding}.  
We define the prototype set as
\[
\mathcal{P} = \{ \mu_c \}_{c=1}^{K},
\]
where \( K = K_{\text{base}} + K_{\text{new}} + 1 \), and each category corresponds to a prototype \( \mu_c \).  
Each prototype lies on the unit hypersphere \( \mathbb{S}^{d-1} \) and is dynamically updated during training.  
The likelihood of a feature \( z_i \) belonging to class \( c \) is modeled by the von Mises–Fisher (vMF) distribution \cite{mardia2009directional}:
\[
p_{\text{vMF}}(z_i;\mu_c,\tau) = C_p(1/\tau)\exp(\mu_c^\top z_i / \tau),\quad c=1,2,\ldots,K,
\]
where \( \tau \) is the temperature parameter.

The concentration parameter is given by $\kappa = 1/\tau,$
and the normalization constant of the vMF distribution is
\[
C_p(\kappa)=\frac{\kappa^{p/2-1}}{(2\pi)^{p/2} I_{p/2-1}(\kappa)},
\]
where \( I_\nu \) denotes the modified Bessel function of the first kind and order \( \nu \), and \( \mu_c \) is the mean direction of the vMF distribution.

The posterior probability that sample \( x_i \) belongs to class \( k \) is
\[
\begin{aligned}
p(y{=}k\,|\,z_i,\tau)
&=
\frac{p_{\mathrm{vMF}}(z_i;\mu_k,\tau)}
     {\sum_{c=1}^{K} p_{\mathrm{vMF}}(z_i;\mu_c,\tau)}
\\[3pt]
&=
\frac{\exp(\mu_k^\top z_i / \tau)}%
     {\sum_{c=1}^{K}\exp(\mu_c^\top z_i / \tau)} .
\end{aligned}
\]

Thus, logits are computed as the similarity between sample features and class prototypes, producing the predictive posterior
\[
p(z_i,\tau)=\left(p(y=1\,|\,z_i,\tau),\ldots,p(y=K\,|\,z_i,\tau)\right)\in\mathbb{R}^{K}.
\]

In this work, we extend prototype learning to the NCD setting and apply prototype modeling to unlabeled data.

\subsubsection{Training Model}
Given an input image, we generate two augmented views 
\((\hat{x}_{i,k}, \tilde{x}_{i,k}, m_{i,k})\) and 
\((\hat{x}'_{i,k}, \tilde{x}'_{i,k}, m'_{i,k})\).
Following DINO~\cite{zhang2022dino}, the two views are fed to a teacher–student network that shares the AMG-ViT backbone, and the teacher produces soft pseudo-labels instead of one-hot targets.

For labeled data \(\mathcal{D}_l\), the teacher receives supervised signals and outputs class-level predictions \(\hat{q}_{i,k}\); for unlabeled data \(\mathcal{D}_u\), it is updated by a momentum objective as in~\cite{chen2020simple}. For each sub-image \(\tilde{x}_{i,k}\) with teacher predictions \(\hat{q}_{i,k}, \tilde{q}_{i,k} \in \mathbb{R}^{N_c+L}\), we use a temperature \(\tau_{\mathrm{sup}}\) to sharpen predictions for known classes, while the student adopts a smoother temperature \(\tau_{\mathrm{stu}}\) to stabilize training.

In industrial anomaly settings, normal regions from different classes often share similar fine-grained appearance, and we assume they share a common normal pseudo-label. In contrast, anomalous regions lack clear semantic cues and may receive incorrect class labels, pushing unknown anomalies toward wrong categories and reducing class discrimination.

To mitigate this, pseudo-labels \(\hat{q}_{i,k}\) are refined using the anomaly score \(s_{i,k}\), enhancing intra-class compactness and suppressing unintended normal-like features in anomalous regions. The refinement is computed as

\[
\hat{q}_{i,k}
\leftarrow 
w_{i,k}\,\mathbf{e} + (1 - w_{i,k}) \hat{q}_{i,k},
\qquad
w_{i,k} = \max(0.5 - s_{i,k},\, 0),
\]
where $\mathbf{e}$ is the one-hot encoding of the normal class.
A lower anomaly score $s_{i,k}$ indicates a higher likelihood of being a normal region; thus, the refined label is pulled closer to $\mathbf{e}$.
The second view $\tilde{q}_{i,k}$ is refined identically.

To ensure consistency between the two augmented views and stabilize training, we assume that predictions for normal regions remain invariant under minor perturbations. Thus, protoAnomalyNCD enforces cross-view prediction consistency for all samples, enabling robust pseudo-label learning. The dual-view consistency loss over a mini-batch \(\mathcal{B}\) of samples is defined as
\[
\mathcal{L}_{\mathrm{dapl}}
=\frac{1}{2|\mathcal{B}|}
\sum_{i\in\mathcal{B}}
\left( 
\ell(q_i, p_i)
+
\ell(\tilde{q}_i, \tilde{p}_i)
\right),
\]
where \(\ell(q,p)=\sum_k -q^{(k)}\log p^{(k)}\) is the cross-entropy.

For labeled data, protoAnomalyNCD directly applies supervised learning to both views:
\[
\mathcal{L}_{\mathrm{sup}}
=\frac{1}{2|\mathcal{B}_l|}
\sum_{i\in\mathcal{B}_l}
\left( 
\ell(y_i, p_i)
+
\ell(y_i, \tilde{p}_i)
\right).
\]

Following~\cite{vaze2022generalized, fei2022xcon}, the unsupervised contrastive loss is
\[
\mathcal{L}_{\mathrm{con}}^{u} =
\frac{1}{|\mathcal{B}|}
\sum_{i\in\mathcal{B}}
-\log
\frac{
\exp(h_i^{T}h_i'/\tau_c)
}{
\sum_{j:K[j\neq i]}\exp(h_i^{T}h_j/\tau_c)
},
\]
where \(K[\cdot]\) is an indicator and \(\tau_c\) is a temperature parameter.

For labeled samples, the supervised contrastive loss is
\[
\mathcal{L}_{\mathrm{con}}^{l}
=\frac{1}{|\mathcal{B}_l|}
\sum_{i\in\mathcal{B}_l}
\frac{1}{|\mathcal{N}(i)|}
\sum_{q\in\mathcal{N}(i)}
-\log 
\frac{
e^{h_i^\top h_q/\tau_c}
}{
\sum_{j\neq i} e^{h_i^\top h_j/\tau_c}
},
\]
where \(\mathcal{N}(i)\) is the set of positive samples sharing the same label as \(x_i\).

To ensure long-term stability in evolving environments, we introduce a marginal entropy maximization term:
\[
\begin{aligned}
\mathcal{L}_{\mathrm{entropy}}
&= -H(\bar{p})
= -\sum_{k=1}^{K} \bar{p}^{(k)} \log \bar{p}^{(k)}, \\
\bar{p}
&= 
\frac{1}{2|\mathcal{B}|}
\sum_{i\in\mathcal{B}}
\left(
p(z_i, \tau_{\mathrm{base}})
+
p(z'_i, \tau_{\mathrm{base}})
\right).
\end{aligned}
\]

This term imposes a roughly uniform prior over categories and acts as a flexible regularizer that adapts to different datasets without extra optimization.

While pseudo-label refinement improves intra-class compactness, classification also benefits from stronger inter-class separation. We therefore explicitly maximize the distance among class prototypes via
\[
\mathcal{L}_{\mathrm{sep}}
=
\frac{1}{K}
\sum_{i=1}^{K}
\log
\frac{1}{K-1}
\sum_{j\neq i}
\exp(\mu_i^{T}\mu_j/\tau_{\mathrm{sep}}),
\]
where \(\tau_{\mathrm{sep}}\) is a temperature hyperparameter.

Combining all objectives, the final loss is
\[
\begin{aligned}
\mathcal{L}_{\mathrm{total}} =\;
&(1-\lambda_{\mathrm{sup}})
\left(
\mathcal{L}_{\mathrm{dapl}}
+
\mathcal{L}_{\mathrm{con}}^{u}
\right)
+
\lambda_{\mathrm{sup}}
\left(
\mathcal{L}_{\mathrm{sup}}
+
\mathcal{L}_{\mathrm{con}}^{l}
\right)
\\
&\quad
+
\lambda_{\mathrm{entropy}}\,\mathcal{L}_{\mathrm{entropy}}
+
\lambda_{\mathrm{sep}}\,\mathcal{L}_{\mathrm{sep}} ,
\end{aligned}
\]
where \(\lambda_{\mathrm{sup}}\in[0,1]\) controls the balance between supervised and unsupervised learning, and \(\lambda_{\mathrm{entropy}},\lambda_{\mathrm{sep}}\) control the strength of the two regularization terms.

\subsection{Estimating the Number of Novel Classes}

In the NCD literature, most methods assume that the number of new classes 
\(K_{\text{new}}\) is known a priori, which is unrealistic in practical 
industrial scenarios.  
Given the complete training dataset 
\(\mathcal{D} = \mathcal{D}_l \cup \mathcal{D}_u\), estimating 
\(K_{\text{new}}\) is therefore essential.  
In this work, we propose an estimation strategy that jointly leverages the 
classification accuracy of labeled samples and the feature statistics of 
labeled data.

Let \(\tilde{K}_{\text{new}}\) denote a candidate estimate.  
ProtoAnomalyNCD evaluates all candidate values using a scoring mechanism based 
on classification accuracy.

We first compute the accuracy score on the base classes:
\[
\text{accScore}
= \frac{1}{|\mathcal{D}_l|}
\sum_{i \in \mathcal{D}_l}
\mathbb{I}\!\left[y_i = \arg\max_c p(y=c \mid z_i, \tau)\right].
\]

When \(\tilde{K}_{\text{new}} > K_{\text{new}}\), many base-class samples in 
\(\mathcal{D}_l\) belonging to the class set \(C_{\text{base}}\) will be 
incorrectly assigned to non-base classes, reducing \text{accScore}.

The feature center of each base class can be computed in two ways:
\[
c^{\,l}_{k}
= \frac{1}{|\mathcal{D}^{\,l}_k|}
\sum_{i \in \mathcal{D}^{\,l}_k} z_i,
\qquad k=1,2,\ldots,K_{\text{base}},
\]
\[
c^{\,u}_{k}
= \frac{1}{|\mathcal{D}^{\,u}_k|}
\sum_{i \in \mathcal{D}^{\,u}_k} z_i,
\qquad k=1,2,\ldots,K_{\text{base}},
\]
where  
\(\mathcal{D}^{\,l}_k\)  
denotes the labeled samples belonging to the base-class label,
and  
\(\mathcal{D}^{\,u}_k\)  
denotes the unlabeled samples assigned to the base class by  
\(\hat{y}_i = \arg\max_c p(y=c \mid z_i, \tau_u)\).

Similarly, under cross-entropy training, when \(\tilde{K}_{\text{new}} < K_{\text{new}}\), many unlabeled samples from novel classes \(C_{\text{new}}\) in \(\mathcal{D}_u\) are incorrectly assigned to base classes \(C_{\text{base}}\). This mismatch enlarges the feature gap between \(c^{\,l}_{k}\) and its corresponding \(c^{\,u}_{k}\), thereby reducing
\[
\text{centrScore}
= \prod_{k=1}^{K_{\text{base}}}
c^{\,l}_{k} c^{\,u}_{k}.
\]

We combine these two metrics to compute the prototype score and select the estimate with the maximum score:
\[
\text{protoScore}(\tilde{K}_{\text{new}})
= \text{accScore} \times \text{centrScore}.
\]

\subsection{Extension to Out-of-Distribution Detection}

In real-world deployment, a model inevitably encounters samples outside the known categories \(C_{\text{base}} \cup C_{\text{new}}\), such as unseen anomaly types or entirely new objects, which we denote as \(C_{\text{out}}\). Forcing these samples into known categories leads to misleading predictions. As illustrated in Fig.~3, a classifier trained only on base and newly discovered classes cannot reject unknown inputs, making out-of-distribution (OOD) detection crucial for reliable industrial inspection.

\begin{figure}[htb]
        \centering
	\includegraphics[width=0.42\textwidth]{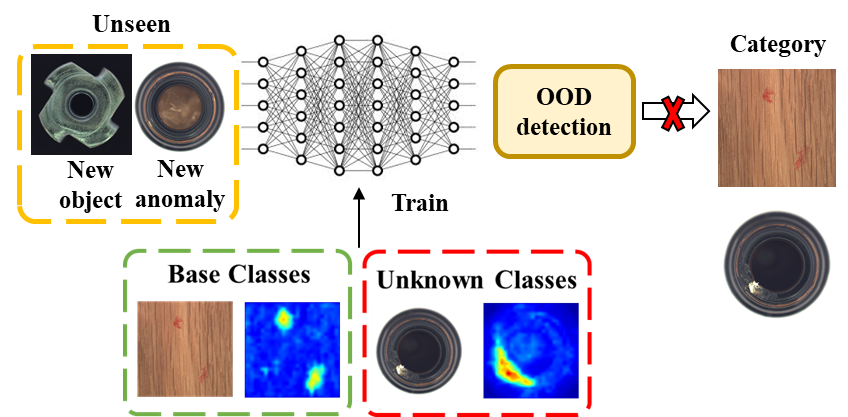}
	\caption{Out-of-Distribution Detection Workflow for Industrial Anomaly Inspection}
	\label{Fig:3}       
\end{figure}

To address this issue, we treat \( C_{\text{base}} \cup C_{\text{new}} \) as the in-distribution (ID) set, while unseen categories are considered OOD. If the model exhibits low confidence for a given input, we regard the sample as potentially OOD and reject its predicted label. Specifically, we compute an OOD score \( S(x) \) and compare it to a threshold \( \delta_{\text{ood}} \); samples with \( S(x) < \delta_{\text{ood}} \) are classified as OOD, and otherwise treated as ID.
Since ProtoAnomalyNCD employs a multi-class classifier, confidence scores
\( p(y=c \mid z) \) are readily available. This enables direct application of post-hoc OOD detection methods such as MSP and Energy without additional training. For example, MSP computes
\[
S(x) = \max_k p(y=k \mid z),
\]
where a lower maximum probability indicates that the sample is unlikely to belong to any known class. Unlike supervised methods that require auxiliary outlier data, these score-based techniques integrate seamlessly with ProtoAnomalyNCD.
Consequently, ProtoAnomalyNCD naturally supports OOD detection during deployment, providing a unified framework that both identifies novel anomaly categories and rejects unseen samples.

\section{Experiments}

\subsection{Experimental Setup}

Experiments are conducted on three industrial anomaly datasets, namely \textbf{MVTec AD}~\cite{bergmann2019mvtec}, \textbf{Real-IAD}~\cite{wang2024real}, and \textbf{Magnetic Tile Defect (MTD)}~\cite{huang2020surface}. MVTec AD contains 10 object and 5 texture categories, each with at least two anomaly types, and the combined anomaly category is removed following \cite{lee2025uniformaly, sohn2023anomaly} for fair comparison. Real-IAD includes 30 object categories captured from five viewpoints, yielding about 150K high-resolution images. MTD consists of 952 normal and 392 anomalous images split into five anomaly types; following \cite{sohn2023anomaly}, 80\% of the normal images are used as reference and the rest for testing. For all datasets, the single-blade subset of the \textbf{Aero-engine Blade Anomaly Detection Dataset (AeBAD-S)}~\cite{zhang2023industrial}, with its normal portion removed, serves as the default labeled image set.

The proposed method is compared with state-of-the-art industrial anomaly clustering approaches, \textbf{AnomalyNCD}~\cite{huang2025anomalyncd} and \textbf{Anomaly Clustering}~\cite{sohn2023anomaly}. Anomaly Clustering is evaluated in two configurations, an unsupervised setting that uses only unlabeled images and a semisupervised setting that additionally uses labeled normal images from the same product, similar to one-class anomaly detection. The comparison further includes deep clustering methods \textbf{UniFormaly}~\cite{lee2025uniformaly}, \textbf{GAT-Cluster}~\cite{niu2020gatcluster}, and \textbf{AMEND}~\cite{banerjee2024amend}, which directly cluster unlabeled images, as well as NCD methods \textbf{GCD}~\cite{vaze2022generalized} and \textbf{SimGCD}~\cite{wen2023parametric}.

Evaluation uses three standard clustering metrics, including F1 score, Normalized Mutual Information (NMI)~\cite{manning2009introduction}, and Adjusted Rand Index (ARI)~\cite{rand1971objective}. Predicted clusters are matched to ground-truth labels with the Hungarian algorithm~\cite{kuhn1955hungarian}. For anomaly map–based methods, multi-class anomaly detection is further assessed using AUPRO at 30\% FPR following \cite{bergmann2019mvtec}.

\subsection{Comparison with State-of-the-Art Methods}
\begin{table*}[h]
\centering
\scriptsize
\setlength{\tabcolsep}{4pt}
\caption{Quantitative results on the MVTec AD, MTD and Real-IAD dataset. All the methods only use unlabeled images as input. }
\begin{tabular}{lcccccccc}
\hline
Dataset & Metric & GATCluster\cite{niu2020gatcluster} & GCD\cite{vaze2022generalized} & SimGCD\cite{wen2023parametric} & AMEND\cite{banerjee2024amend} & AC\cite{sohn2023anomaly} & AnomalyNCD\cite{huang2025anomalyncd} & \shortstack{INPformer\cite{luo2025exploring} \\ + ours} \\
\hline
MVTec & NMI & 0.136 & 0.417 & 0.452 & 0.431 & 0.525 & 0.613 & 0.647 \\
      & ARI & 0.053 & 0.302 & 0.346 & 0.333 & 0.431 & 0.526 & 0.582 \\
      & F1  & 0.264 & 0.553 & 0.569 & 0.542 & 0.604 & 0.712 & 0.744 \\
\hline
MTD   & NMI & 0.028 & 0.211 & 0.105 & 0.138 & 0.179 & 0.268 & 0.343 \\
      & ARI & 0.009 & 0.115 & 0.048 & 0.067 & 0.120 & 0.228 & 0.281 \\
      & F1  & 0.243 & 0.381 & 0.293 & 0.324 & 0.346 & 0.509 & 0.522 \\
\hline
Real-IAD & NMI & 0.102 & 0.116 & 0.152 & 0.131 & 0.323 & 0.381 & 0.423 \\
         & ARI & 0.131 & 0.157 & 0.201 & 0.189 & 0.301 & 0.377 & 0.397 \\
         & F1  & 0.263 & 0.552 & 0.561 & 0.529 & 0.481 & 0.601 & 0.659 \\
\hline
\end{tabular}
\end{table*}

\begin{table*}[h]
\centering
\scriptsize
\setlength{\tabcolsep}{4pt}
\caption{ Quantitative results on the MVTec AD, MTD and Real-IAD dataset. All the methods use unlabeled images and labeled normal images as input.}
\begin{tabular}{lcccccccc}
\hline
Dataset & Metric & AnomalyNCD\cite{huang2025anomalyncd} & AC\cite{sohn2023anomaly} & UniFormaly\cite{lee2025uniformaly} &
\shortstack{EffAD\cite{batzner2024efficientad} \\ + ours} &
\shortstack{PatchCore\cite{roth2022towards} \\ + ours} &
\shortstack{R++\cite{tien2023revisiting} \\ + ours} &
\shortstack{INPformer\cite{luo2025exploring} \\ + ours} \\
\hline
MVTec & NMI & 0.631 & 0.608 & 0.547 & 0.591 & 0.683 & 0.701 & 0.758 \\
      & ARI & 0.542 & 0.489 & 0.433 & 0.505 & 0.665 & 0.687 & 0.721 \\
      & F1  & 0.721 & 0.652 & 0.645 & 0.692 & 0.791 & 0.821 & 0.856 \\
\hline
MTD   & NMI & 0.368 & 0.391 & 0.421 & 0.302 & 0.346 & 0.359 & 0.381 \\
      & ARI & 0.361 & 0.314 & 0.322 & 0.251 & 0.381 & 0.389 & 0.402 \\
      & F1  & 0.601 & 0.491 & 0.609 & 0.439 & 0.528 & 0.582 & 0.620 \\
\hline
Real-IAD & NMI & 0.415 & 0.373 & 0.391 & 0.337 & 0.429 & 0.447 & 0.491 \\
         & ARI & 0.407 & 0.325 & 0.372 & 0.329 & 0.449 & 0.451 & 0.468 \\
         & F1  & 0.621 & 0.532 & 0.529 & 0.531 & 0.693 & 0.718 & 0.757 \\
\hline
\end{tabular}
\end{table*}

In Table~1, all methods cluster only unlabeled images from MVTec AD, MTD, and Real-IAD. ProtoAnomalyNCD combined with the prototype-based AD method INP-former~\cite{luo2025exploring} consistently improves clustering quality on all three datasets. These results indicate that the proposed contrastive learning framework captures more discriminative anomaly features and encourages the model to focus on anomaly-relevant regions.

In Table~2, all methods use both unlabeled images and labeled normal (base-class) images from the same product. ProtoAnomalyNCD is integrated with several AD backbones. Among these variants, INP-former achieves the best AUPRO and the strongest overall clustering performance. These observations show that leveraging labeled base-class images further enhances the discriminative power of ProtoAnomalyNCD and leads to more accurate multi-class anomaly clustering. Furthermore, the performance differences across AD methods indicate that the selection of the front-end anomaly detection backbone significantly influences the overall effectiveness of the framework.

Previous anomaly detection works mostly assume binary detection or an already known number of categories. 
To relax this constraint, we propose a \textbf{dynamic Prototype Score} for class number estimation. 
We compare with GCD~\cite{vaze2022generalized} and SimGCD~\cite{wen2023parametric}; results are shown in Table~3.

\begin{table}[h]
\centering

\caption{Category number estimation.}
\begin{tabular}{lcccc}
\hline
Dataset & Ground Truth & GCD & SimGCD & Ours \\
\hline
MVTec & 84 & 53 & 56 & 69 \\
Real-IAD & 138 & 76 & 82 & 115 \\
\hline
\end{tabular}
\end{table}

\subsection{Ablation Studies}

\begin{figure}[htb]
        \centering
	\includegraphics[width=0.28\textwidth]{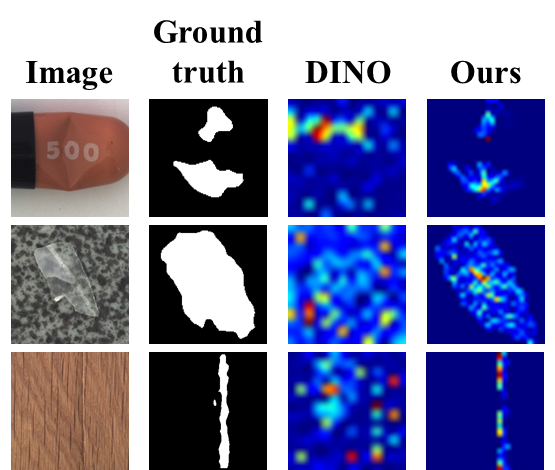}
	\caption{Visualization of the self-attention of the [CLS] token
    on the last layer’s heads. DINO attention refers to the [CLS]
    token extracted from a DINO pre-trained ViT that mainly focuses
    on a foreground object. ProtoAnomalyNCD uses a anomaly map to direct the
    [CLS] token’s attention to the anomalous regions.}
	\label{Fig:4}       
\end{figure}

The proposed Anomaly Map Guided Attention encourages the ViT to assign high responses to true anomaly regions while suppressing background and normal object areas. Figure~\ref{Fig:4} compares anomaly maps predicted by a vanilla DINO-pretrained ViT and by our AMGA-enhanced model on several MVTec AD samples. The baseline maps are diffuse and often focus on large object areas, whereas our maps align much better with the ground-truth anomaly masks.

As shown in Table~4, AMGA also brings consistent quantitative improvements. Compared with the vanilla DINO baseline, our method improves NMI by 4.6\%, ARI by 3.3\%, and F1 by 2.4\%, respectively, confirming the effectiveness of anomaly-map-guided attention.

\begin{table}[h]
\centering

\caption{AMGA effectiveness on the MVTec AD.}
\begin{tabular}{lccc}
\hline
Metric & DINO & All Tokens & Ours \\
\hline
NMI & 0.601 & 0.539 & 0.647 \\
ARI & 0.549 & 0.472 & 0.582 \\
F1  & 0.720 & 0.651 & 0.744 \\
\hline
\end{tabular}
\end{table}

\begin{figure}[htb]
        \centering
	\includegraphics[width=0.47\textwidth]{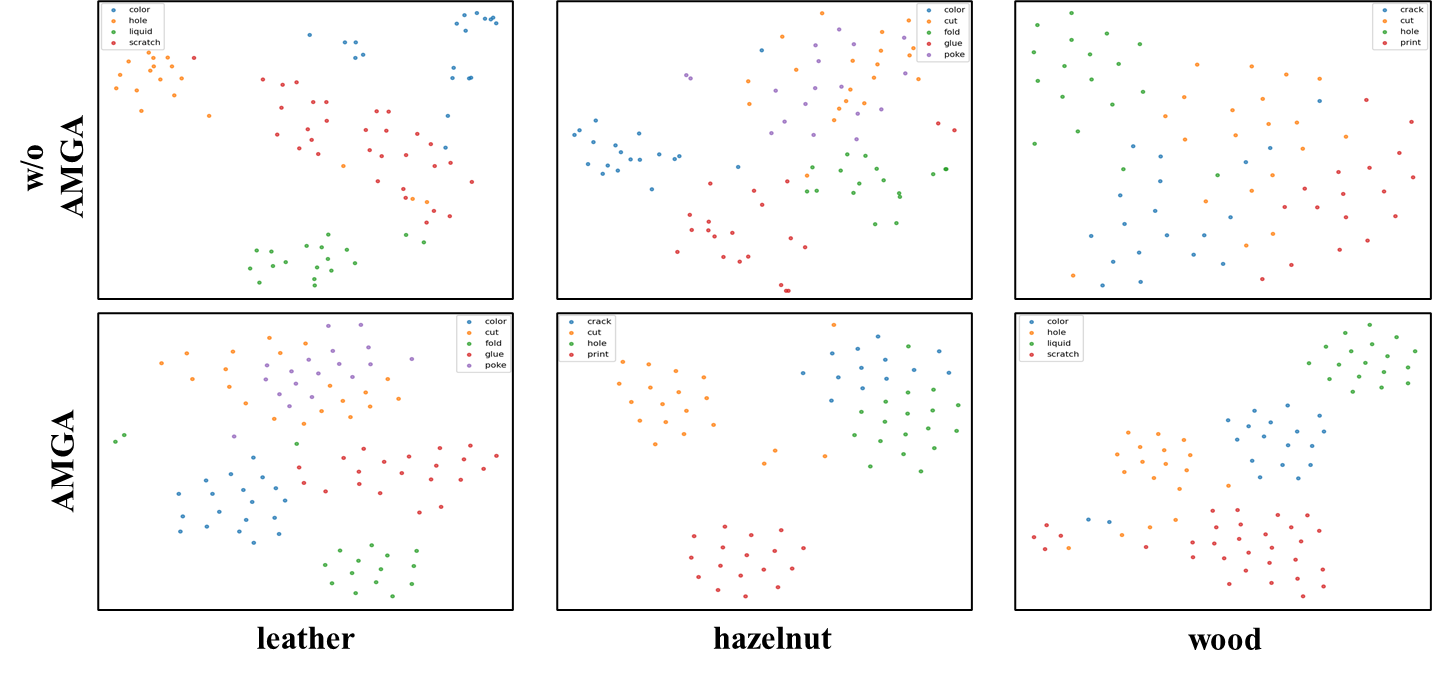}
	\caption{T-SNE visualization of sub-images on the MVTec AD
dataset. We choose leather, hazelnut and wood as examples.
The different colors of dots represent their anomaly classes.}
	\label{Fig:5}       
\end{figure}

Figure~\ref{Fig:5} presents t-SNE visualizations of anomaly features for three MVTec AD categories (leather, hazelnut, and wood), with and without AMGA. After applying AMGA, the embeddings form tighter intra-class clusters and clearer inter-class boundaries, indicating more discriminative anomaly representations.

\begin{table}[h]
\centering
\caption{Effect of prototype learning.}
\begin{tabular}{lcccc}
\hline
Dataset & Metric & KMeans & w/o Proto & Ours \\
\hline
MVTec & NMI & 0.604 & 0.621 & 0.647 \\
      & ARI & 0.518 & 0.534 & 0.582 \\
      & F1  & 0.715 & 0.729 & 0.744 \\
\hline
MTD   & NMI & 0.219 & 0.283 & 0.343 \\
      & ARI & 0.157 & 0.238 & 0.281 \\
      & F1  & 0.422 & 0.492 & 0.522 \\
\hline
\end{tabular}
\end{table}

To assess the role of prototypes, we freeze AMG-ViT and compare ProtoAnomalyNCD with a standard classification head and k-means clustering. As shown in Table~5, a single prototype suffices to represent each base or novel category, and the model pulls samples toward their category-level prototype. This stabilizes clustering of unlabeled categories, prevents collapse among novel classes, and yields a more separable embedding space and more balanced accuracy across base and novel anomaly classes.

Table~6 ablates the main loss terms of ProtoAnomalyNCD. Removing the unsupervised contrastive loss \(L_{\text{unsup}}\) weakens structural learning on unlabeled images and degrades new-class clustering. Without cross-view alignment \(L_{\text{dapl}}\), teacher–student predictions drift and pseudo labels deteriorate. Dropping marginal entropy \(L_{\text{Memax}}\) causes cluster imbalance and collapse, while removing prototype separation \(L_{\text{sep}}\) reduces inter-prototype margins and harms class separability. These results show that all components contribute in a complementary way.

\begin{table}[h]
\centering

\setlength{\tabcolsep}{4pt}
\caption{Ablation on major losses.}
\begin{tabular}{lcccccc}
\hline
Dataset & Metric & \shortstack{w/o \\ $L_{\text{dapl}}$} & \shortstack{w/o \\ $L_{\text{sup}}$} & \shortstack{w/o \\ $L_{\text{Memax}}$} & \shortstack{w/o \\$L_{\text{sep}}$} & Ours \\
\hline
MVTec & NMI & 0.296 & 0.577 & 0.598 & 0.582 & 0.647 \\
      & ARI & 0.261 & 0.519 & 0.526 & 0.525 & 0.582 \\
      & F1  & 0.460 & 0.704 & 0.711 & 0.706 & 0.744 \\
\hline
MTD   & NMI & 0.094 & 0.258 & 0.296 & 0.192 & 0.343 \\
      & ARI & 0.104 & 0.203 & 0.244 & 0.175 & 0.281 \\
      & F1  & 0.379 & 0.416 & 0.501 & 0.419 & 0.522 \\
\hline
\end{tabular}
\end{table}

Our PLC mitigates over-detection in anomaly localization, where visually diverse normal patches tend to receive inconsistent pseudo labels. As reported in Table~7, applying PLC improves NMI, ARI, and F1 by 2.3\%, 1.6\%, and 0.5\%, respectively, leading to more stable clustering and slightly better overall multi-class anomaly classification.

\begin{table}[h]
\centering

\caption{The ablation experiment of pseudo label correction (PLC) on the MVTec AD dataset.}
\begin{tabular}{lccc}
\hline
 Metric & w/o PLC & Ours \\
\hline
 NMI & 0.624 & 0.647 \\
 ARI & 0.566 & 0.582 \\
 F1  & 0.739 & 0.744 \\
\hline
\end{tabular}
\end{table}

For the OOD detection scenario in Sec.~3.5, we randomly sample 20\% of samples outside each category as OOD inputs. 
Following \cite{hendrycks2016baseline,hendrycks2019scaling}, we evaluate AUROC and FPR95, treating ID classes 
$C_{\text{base}} \cup C_{\text{new}}$ 
as positive and OOD classes 
$C_{\text{out}}$ 
as negative. 
We test ProtoAnomalyNCD under multiple scoring functions, including 
\textbf{Max Logit Score (MLS)}~\cite{hendrycks2019scaling} 
and 
\textbf{Energy}~\cite{liu2020energy}. 
MLS explores logits via similarity to prototypes, whereas Energy better follows density variations in feature space. 
As shown in Table~8, both MLS and Energy outperform the MSP baseline.

\begin{table}[h]
\centering

\caption{OOD detection performance.}
\begin{tabular}{lcc|cc}
\hline
Method & \multicolumn{2}{c|}{MVTec} & \multicolumn{2}{c}{Real-IAD} \\
       & FPR95 & AUROC & FPR95 & AUROC \\
\hline
MSP\cite{hendrycks2016baseline}    & 73.90 & 44.71 & 69.12 & 51.98 \\
MLS\cite{hendrycks2019scaling}     & 74.20 & 45.13 & 68.21 & 52.02 \\
Energy\cite{liu2020energy} & 75.90 & 45.01 & 70.45 & 53.17 \\
\hline
\end{tabular}
\end{table}

\section{Conclusion}
We propose ProtoAnomalyNCD, a prototype-based framework for multi-class industrial anomaly classification that is compatible with existing anomaly detection methods and serves as an initial step toward generalized semantic anomaly analysis. Grounded SAM with text prompts is used to localize object regions and provide strong priors for the anomaly classifier, and an anomaly-mask-guided attention mechanism leverages localized product regions and anomaly maps as region-guidance priors for prototype learning. Within this unified prototype-learning framework, ProtoAnomalyNCD discovers and clusters novel anomalies while simultaneously performing multi-type anomaly classification, and can be further extended to detect unseen out-of-distribution anomalies, thereby unifying anomaly discovery, classification, and rejection. Although its performance may be affected by the quality of the underlying anomaly detector, ProtoAnomalyNCD still substantially outperforms existing industrial multi-class anomaly classification methods, and we hope it will inspire further research on versatile, open-set industrial anomaly analysis under more realistic conditions.

{
    \small
    \bibliographystyle{ieeetr}
    \bibliography{main}
}


\end{document}